\pdfoutput=1

\documentclass[11pt]{article}

\usepackage[final]{acl}

\usepackage{times}
\usepackage{latexsym}

\usepackage[T1]{fontenc}

\usepackage[utf8]{inputenc}

\usepackage{microtype}

\usepackage{inconsolata}

\usepackage{pmboxdraw}
\usepackage{subfig}
\usepackage{graphicx}
\usepackage{multirow}
\usepackage{times}
\usepackage{latexsym}
\usepackage{amssymb}
\usepackage{amsmath}
\usepackage{amsthm}
\usepackage{booktabs}
\usepackage{algorithm}
\usepackage{algorithmic}
\usepackage{color}
\usepackage{colortbl}

\definecolor{mred}{RGB}{236,151,165}
\definecolor{mgreen}{RGB}{200,229,179}
\definecolor{myellow}{RGB}{252,241,210}
\definecolor{mblue}{RGB}{65,102,245}

\newtheorem{observation}{Observation}
\newtheorem{hypothesis}{Hypothesis}

\title{A Text is Worth Several Tokens:\\Text Embedding from LLMs Secretly Aligns Well with The Key Tokens}

\author{
    Zhijie Nie\textsuperscript{\rm 1,3}, Richong Zhang\textsuperscript{\rm 1,2}\thanks{\ \ Corresponding author}, Zhanyu Wu\textsuperscript{\rm 1}\\
    \textsuperscript{\rm 1}CCSE, School of Computer Science and Engineering, Beihang University, Beijing, China\\
    \textsuperscript{\rm 2}Zhongguancun Laboratory, Beijing, China\\
    \textsuperscript{\rm 3}Shen Yuan Honors College, Beihang University, Beijing, China\\
    \texttt{\{niezj,zhangrc,wuzy24\}@act.buaa.edu.cn}
}

\begin{document}
\maketitle
\begin{abstract}
Text embeddings from large language models (LLMs) have achieved excellent results in tasks such as information retrieval, semantic textual similarity, etc. In this work, we show an interesting finding: when feeding a text into the LLM-based embedder, the obtained text embedding can be aligned with the key tokens in the input text. We first fully analyze this phenomenon on eight LLM-based embedders and show that this phenomenon is universal and is not affected by model architecture, training strategy, and embedding method. Upon further analysis, we find that the main change in embedding space between these embedders and their LLM backbones lies in the first principal component. By adjusting the first principal component, we can align text embedding with the key tokens. Finally, we demonstrate the broad application potential of this finding: (1) we propose a simple and practical sparse retrieval method based on the aligned tokens, which can achieve 80\% of the dense retrieval effect of the same model while reducing the computation significantly; (2) we show that our findings provide a novel perspective to help understand novel technologies (e.g., instruction-following embedding) and fuzzy concepts (e.g., semantic relatedness vs similarity) in this field\footnote{Our code is available at \url{https://github.com/Arthurizijar/Text_aligns_tokens}}. 

\end{abstract}

\section{Introduction}

Large language models (LLMs) have recently made rapid progress on various natural language understanding tasks using the generative paradigm \cite{brown2020language}. However, not all tasks lend themselves to the generative paradigm in practice; tasks such as information retrieval, text clustering, and semantic text similarity usually rely on high-quality text embeddings. Thus, more and more attention has been focused on obtaining high-quality textual embeddings from large language models \cite{jiang2023scaling, springer2024repetition, behnamghader2024llm2vec}.

\begin{figure}[t]
    \centering
    \includegraphics[width=\linewidth]{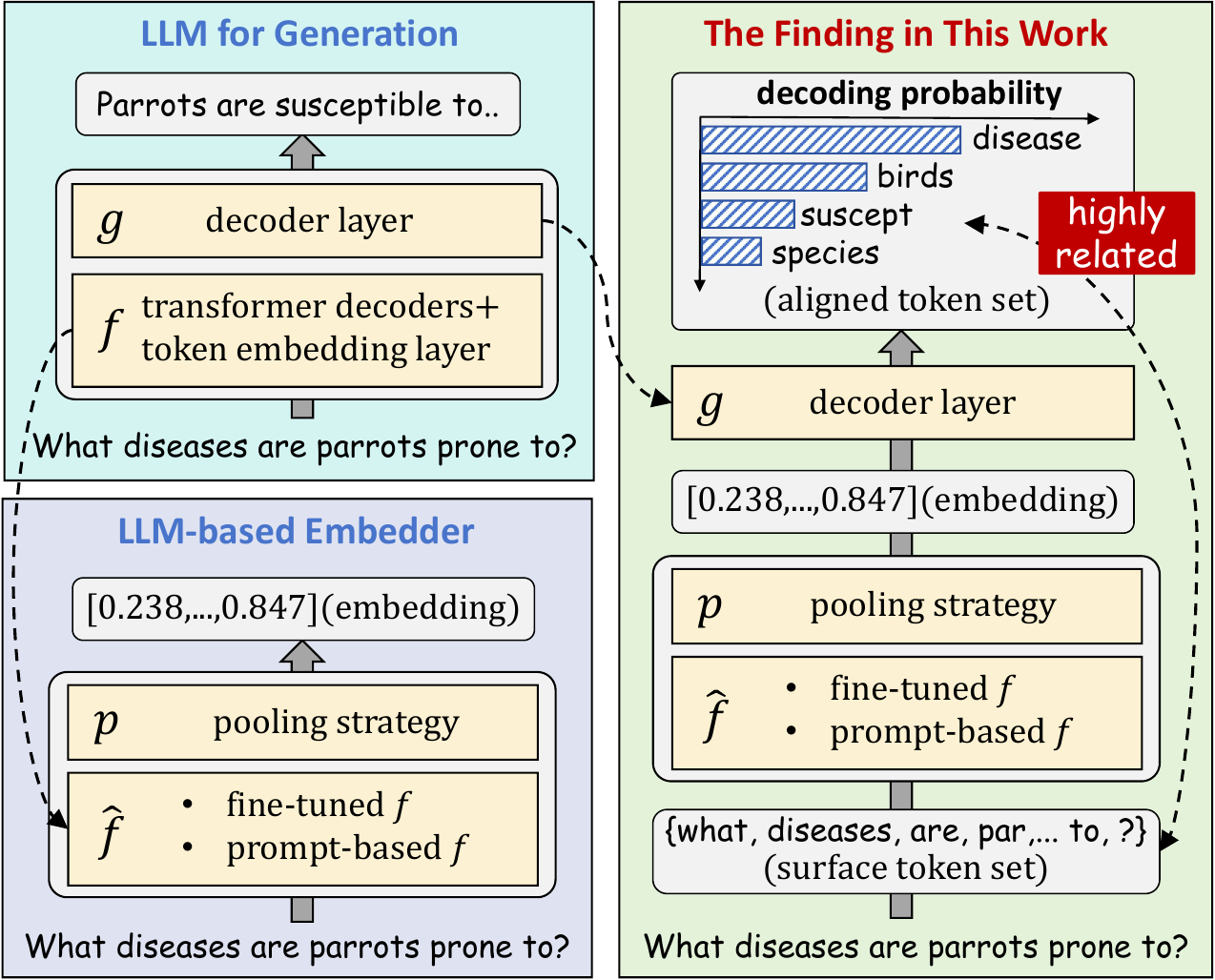}
    \caption{Paradigms on LLMs for text generation and embedding (left) and our novel findings (right). }
    \label{fig:introdution}
\end{figure}

As shown on the left half of Figure \ref{fig:introdution}, the LLM for generation takes the texts as input and output. The input text is tokenized and passed through the module $f$ to obtain its hidden states. Then, a decoder layer $g$ is required, which maps the high-dimensional hidden states to the vocabulary-length logits and computes the decoded probability for each token. When LLMs are converted for text embedding, current methods typically incorporate the following changes: (1) $g$ is discarded because there is no need to map to the vocabulary; (2) $f$ is converted into $\hat{f}$ using prompt-engineering \cite{jiang2023scaling,springer2024repetition} or contrastive learning \cite{muennighoff2022sgpt,behnamghader2024llm2vec}; and (3) a pooling strategy $p$ is used to weight sum of hidden states and obtain the text embedding.

In this paper, we are not proposing a new text embedding method for LLMs. Instead, our research centers on a very interesting finding: when the text embedding obtained by $\hat{f}$ passes through the decoder layer $g$ from the same LLM, the tokens with the highest decoding probability are highly related to the input text. In other words, the embedding of the input text is aligned with some key tokens of that text. As shown in the right half of Figure \ref{fig:introdution}, when the input text is ``\textit{What diseases are parrots prone to ?}'', we can find the literally-related tokens, such as ``{\it disease}'' and the semantically-related tokens, such as ``{\it birds}'' and ``{\it suscept}'' have the highest decoding probabilities.

This phenomenon may not be surprising in some prompt-based methods, which direct LLMs to summarise the whole text in a word (See \S\ref{sec:background_prompt} for details). However, based on the sufficient study of eight LLM-based embedders \footnote{We use ``embedder'' instead of ``encoder'' to prevent unnecessary misunderstanding since the backbones of the current methods are usually decoder-only LLMs.}, we observe that the above phenomenon is universal, independent of the LLMs' architecture, the training strategy, and the embedding method. (\S\ref{sec:align}). Especially this phenomenon appears even more clearly in those methods based on contrastive learning, uncovering the unity among different methods. 

Considering the unusual consistency of this phenomenon, we perform deeper analyses based on these LLMs to understand this finding more precisely. Specifically, we compare the embedding spaces of $f$ and $\hat{f}$ using spectral analysis (\S\ref{sec:analysis}). We find that the dominant change in $\hat{f}$ is concentrated in the first principal component. By manually adjusting the first principal component of the embedding space, we can replicate the phenomenon of aligning text embeddings to key tokens.

With a deeper understanding of our findings, we believe that it has a rich potential for application (\S\ref{sec:application}).  For example, we find that the criticism of LLM-generated embedding mainly stems from its high dimensionality, resulting in significant inference and storage overhead \cite{muennighoff2024generative}. To address this, we propose a new sparse retrieval method based on our findings. We convert document embeddings into a sparse representation consisting only of aligned tokens and utilize a few aligned tokens from the query embedding for expansion. Despite its simplicity, our method achieves over 80\% of the performance of the original LLM-based embedder. At the same time, we show that our work helps to intuitively understand (1) the working mechanism of the instruction-following embedding \cite{su2023one} and (2) the influence of training data on the embedding space.

Our contributions are summarized as follows:
\begin{itemize}
    \item We find that the text embeddings obtained in the LLM-based embedders align with the key tokens, providing a unified perspective for understanding prompt engineering methods and contrastive learning methods; 
    \item We explain why this phenomenon occurs from the perspective of spectral analysis and find that the current method mainly changes the first principal component of the original embedding space of the LLMs;
    \item We show a series of application examples, including improvements to the method and interpretability of the model, demonstrating the large application potential of our findings.
\end{itemize}

\section{Background}\label{sec:background}
\subsection{Basic Paradigm}
Given a LLM $F$, we can divide it into two parts:
\begin{equation}
    F=g \circ f
\end{equation}
where $g$ is the decoder layer, and $f$ is the rest modules of the LLM. In the existing LLM embedding methods, $g$ is discarded, while $f$ can be used as a text embedder. Given a text $s_i$, we convert it to a token sequence using LLM's tokenizer and get $s_i = \{t_{i1}, \cdots, t_{il}\}$, where $l$ is the sequence length; then we can get the hidden state of the last layer:
\begin{equation}
    {\bf H} = [{\bf h}_{i1}^{(t)},\cdots,{\bf h}_{il}^{(t)}] = f(s_i)
\end{equation}
where ${\bf H} \in \mathbb{R}^{d\times l}$ and ${\bf h}_{ij}^{(t)} \in \mathbb{R}^{d\times 1}$ is the $i$-th $d$-dimensional hidden state. Subsequently, the pooling strategy $p(.)$ is used to $\bf{H}$ for the text embedding ${\bf h}_i$, which can be expressed as
\begin{equation}
    {\bf h}_i = p(f(s_i)) = p({\bf H}) = \sum\nolimits_{j=1}^l \alpha_j {\bf h}_{ij}^{(t)}
\end{equation}
where $\{\alpha_j\}_{j=1}^l$ is the weight factor satisfying $\sum_{j=1}^l \alpha_j = 1$. Specifically, there are three popular pooling strategies in practice: for last pooling, $\alpha_j$ is $1$ if $j=l$ else is $0$; for mean pooling, $\alpha_j=1/l$ for each $j$; for weighted mean pooling \cite{muennighoff2022sgpt}, $\alpha_j=j/\sum\nolimits_{j=1}^l j$.

However, text embeddings obtained directly from the encoder $f$ show poor performance. It is unsurprising since the pre-training task, next token prediction, is not designed for embedding, and the unidirectional attention detracts from the expressive power of the hidden states \cite{li2024bellm}. In the subsequent subsections, we introduce how the existing methods improve the embedding's quality based on the top of $f$. For simplicity, we indiscriminately refer to the LLM-based embedder improved based on $f$ as $\hat{f}$.

\subsection{Embedding via Prompt Engineering}\label{sec:background_prompt}
The embedder $\hat{f}$ based on prompt engineering fills the text into prompt templates to improve the quality of text embedding, which can be expressed as
\begin{equation}
    \hat{f}(s_i)=f(t(s_i))
\end{equation}
where $t(.)$ represents the operation of filling the text into a fixed prompt template.

PromptEOL \cite{jiang2023scaling} introduces a prompt template: \textcolor{mblue}{\texttt{This sentence:"[text]" means in one word:"}}, where \texttt{[text]} is a placeholder. In practice, the template where \texttt{[text]} is replaced by a specific text is sent into the encoder $f$, and the last pooling strategy is used to obtain the text embedding. The following works design a better prompt template based on task-oriented \cite{lei2024meta} or chain-of-thought \cite{zhang2024simple} can lead to better performance.

The methods based on prompt engineering are simple and training-free, so they are unlikely to compromise the LLMs' generation capabilities. However, they provide limited performance improvement for downstream tasks.

\subsection{Embedding via Contrastive Learning}
The methods based on contrastive learning inherited the valuable experience of the BERT-based encoder era \cite{gao2021simcse}. In these methods, $\hat{f}$ is fine-tuned $f$ with contrastive learning. Due to the large parameter count of $f$ itself, parameter-efficient fine-tuning methods such as LoRA \cite{hu2021lora} are usually used.

Given a text dataset $D$, for any text $s_i \in D$, we first obtain its embedding ${\bf h}_i$ from $f$ with a specific pooling strategy. Then positive pairs $({\bf h}_i, {\bf h}_i^+)$ and negative pairs $\{({\bf h}_i, {\bf h}_{ij}^-)\}_{j=1}^N$ are constructed following different settings, where $N$ is the negative example number. In the unsupervised setting, two data-augmented views of a text are considered a positive pair, while the negative samples are randomly sampled from the datasets. In the supervised setting, the positive pair is a labeled text pair, which can be query-document, question-answer or hypothesis-entailment, while hard negative pairs may be introduced. Finally, the contrastive loss can be expressed as
\begin{equation}\label{eqn:cl}
    \mathcal{L}_{\rm cl} = -\log \frac{e^{d({\bf h}_i,{\bf h}_i^+)/\tau}}{e^{d({\bf h}_i,{\bf h}_i^+)/\tau}+\sum_{j=1}^N e^{d({\bf h}_i, {\bf h}_{ij}^-)/\tau}}
\end{equation}
where $d(\cdot,\cdot)$ is a distance function, $\tau$ is the temperature hyper-parameter. During fine-tuning, the contrastive loss draws positive text pairs close while pushing negative text pairs away.

\paragraph{Additional Tricks} There are some effective tricks in the existing works, which include: (1) switching casual attention to bi-directional attention \cite{behnamghader2024llm2vec}; (2) using different instruction prefixes for the datasets from different tasks to minimize inter-task interference \cite{su2023one}; (3) co-training contrastive learning and next word prediction to minimize reductions to generative capability \cite{muennighoff2024generative}.

\begin{table*}[th]
\footnotesize
\begin{tabular}{ccccccc}
\toprule
\multirow{2}{*}{\textbf{Model}} & \multicolumn{2}{c}{\bf Architecture} & \multicolumn{2}{c}{\bf Fine-tuning} &  \multicolumn{2}{c}{\bf Embedding} \\
& \textbf{Backbone} & \textbf{Attention} & \textbf{Paradigm} & \textbf{Corpus} & \textbf{Pooling} & \textbf{Similarity} \\
\midrule
SGPT$_{\text{nli}}$ & GPT-Neo & casual & SCL & NLI & weighted mean & cosine \\
SGPT$_{\text{msmarco}}$ & (1.3B) & casual & SCL & MS MARCO & weighted mean & cosine \\
\midrule
OPT$_{\text{EOL}}$ & OPT & casual & PE & - & last token & dot product \\
OPT$_{\text{EOL+CSE}}$ & (1.3B) & casual & PE+SCL & NLI & last & dot product \\
\midrule
LLaMA$_{\text{EOL}}$ & LLaMA & casual & PE & - & last token & dot product \\
LLaMA$_{\text{EOL+CSE}}$ & (7B) & casual & PE+SCL & NLI & last & dot product \\
\midrule
GritLM & Mistral & bi-directional & SCL+NTP & Tulu 2+E5+S2ORC & mean & cosine \\
LLM2Vec & (7B) & bi-directional & MNTP$\rightarrow$SCL & E5 & weighted mean & cosine \\
\bottomrule \\
\end{tabular}
\caption{Detailed information on the model used to study the embedding space. The paradigms are shortened as follows: supervised contrastive learning (SCL), unsupervised contrastive learning (UCL), prompt engineering (PE), next token prediction (NTP), and masked next token prediction (MNTP) \cite{behnamghader2024llm2vec} separately.}
\label{tab:model}
\end{table*}

\section{Embedding Aligns with Key Tokens}\label{sec:align}

\subsection{Motivation}
To analyze the pre-trained transformer in the embedding space, \citet{elhage2021mathematical,geva2022transformer,dar2022analyzing} attempt to multiply the attention or feed-forward layer parameters with the token embedding matrix to explain how these parameters work. For example, \citet{geva2022transformer} multiplies the feed-forward value vector with the token embedding matrix to obtain a distribution over the vocabulary and find that the tokens with high probability can explain what FFNs update to hidden layer representations. Inspired by these works, we try to interpret text embeddings obtained from LLMs by mapping them into the token space. 

\subsection{Method}
To implement the above idea, we introduce a text dataset $D$, and a triplet $(\hat{f}, T, {\bf E}_g)$: $\hat{f}$ is the LLM-based embedder, $T=\{t_1,\cdots,t_L\}$ is the $L$-sized vocabulary and ${\bf E}_g = [{\bf e}_{t_1},\cdots,{\bf e}_{t_L}] \in \mathbb{R}^{d\times L}$ is the token embedding matrix from the decoded layer $g$, where ${\bf e}_{t_j} \in \mathbb{R}^{d \times 1}$ is the token embedding of token $t_j$. Note that $T$ and ${\bf E}_g$ are determined by the original LLM $F$ and ${\bf E}_g$ is the only parameter in $g$~\footnote{To the best of our knowledge, all popular LLMs follow the original design of the decoder layer from GPT \cite{radford2018improving}, i.e., a linear layer without bias, which also can be regarded as a token embedding matrix. }, therefore, there is no difference between ${\bf E}_g^\top {\bf h}_i$ and $g({\bf h}_i)$ for any text embedding ${\bf h}_i \in \mathbb{R}^{d\times 1}$.

Given a text $s_i \in D$, we obtain its literal token set $T_{s_i}$ and top $K$ aligned token set $\hat{T}_{s_i}^K$ then capture the potential connection between these two sets. For $T_{s_i}$, we (1) convert $s_i$ into tokens by the tokenizer of $f$ and (2) deduplicate the token sequence to form a token set $T_{s_i}$. For $\hat{T}_{s_i}^K$, we (1) follow the pooling strategy of $\hat{f}$ to obtain the text embedding ${\bf h}_i$, (2) calculate the dot product between ${\bf h}_i$ and the token embedding ${\bf e}_{t_j}$ for each token $t_j$, (3) obtain the ordered token set $\hat{T}_{s_i}$ by sorting in descending order according to dot-product results, and (4) select the first $K$ elements from $\hat{T}_{s_i}$ to form $\hat{T}_{s_i}^K$. We provide an algorithmic form to describe this process precisely:

\begin{algorithm}[hb]\small
    \renewcommand{\algorithmicrequire}{\textbf{Input:}}
    \renewcommand{\algorithmicensure}{\textbf{Output:}}
    \caption{\small Embedding-Token Alignment Analysis}
    \label{alg:process}
    \begin{algorithmic}[1]
        \REQUIRE A text dataset $D$ and the triplet $(\hat{f}, T, {\bf E}_g)$.
        \STATE Initialization: $i \leftarrow 0$, $j \leftarrow 0$
        \WHILE{$i \leq |D|$}
            \STATE Get the $i$-th text $s_i$ in $D$
            \STATE Deduplicate $\text{tokenizer}(s_i)$ to obtain $T_{s_i}$
            \STATE Calculate ${\bf h}_i \leftarrow \text{pooling}(\hat{f}(s_i))$
            \WHILE{$j \leq |T|$}
                \STATE Calculate ${\rm score}(t_j, s_i) \leftarrow {\bf e}_{t_j}^\top {\bf h}_i$
                \STATE Update $j \leftarrow j + 1$
            \ENDWHILE
            \STATE Sort $T$ in descending by ${\rm score}(t_j, s_i)$ to get $\hat{T}_{s_i}$
            \STATE Select the first $K$ elements from $\hat{T}_{s_i}$ to form $\hat{T}_{s_i}^K$
            \STATE Update $i \leftarrow i + 1$
        \ENDWHILE
        \ENSURE  $T_{s_i}$ and $\hat{T}_{s_i}^K$
    \end{algorithmic} 
\end{algorithm}

\begin{table*}[th]\footnotesize
\centering
\begin{tabular}{l|l}
\toprule
\textbf{Model} & \textbf{Top 10 Aligned Tokens} \\
\midrule
GPT-Neo & \colorbox{mgreen}{\_and} \colorbox{mgreen}{,} \colorbox{mred}{Ċ} \colorbox{mgreen}{\_in} \colorbox{mred}{\_(} \colorbox{mgreen}{.} \colorbox{mred}{\_the} \colorbox{mred}{\_as} \colorbox{mred}{\_on} \colorbox{mred}{\_for}\\ 
SGPT$_{\text{nli}}$ & \colorbox{mgreen}{\_2003} \colorbox{myellow}{2003} \colorbox{mred}{\_03} \colorbox{mred}{\_3} \colorbox{mgreen}{\_March} \colorbox{myellow}{\_game} \colorbox{mgreen}{\_released} \colorbox{mred}{\_three} \colorbox{myellow}{\_games} \colorbox{mred}{03}\\ 
SGPT$_{\text{msmarco}}$ & \colorbox{mgreen}{\_Advance} \colorbox{mgreen}{\_Game} \colorbox{myellow}{\_Released} \colorbox{myellow}{\_Releases} \colorbox{mred}{\_ADV} \colorbox{myellow}{Game} \colorbox{myellow}{\_GAME} \colorbox{myellow}{\_release} \colorbox{mgreen}{\_released} \colorbox{myellow}{\_releases}\\ 
\midrule 
OPT & \colorbox{mred}{Ċ} \colorbox{mred}{\_The} \colorbox{myellow}{\_It} \colorbox{mred}{\_A} \colorbox{myellow}{\_In} \colorbox{mred}{\_This} \colorbox{mgreen}{</s>} \colorbox{mred}{\_An} \colorbox{mred}{\_As} \colorbox{myellow}{\_Its}\\ 
OPT$_{\text{EOL}}$ & \colorbox{myellow}{released} \colorbox{mgreen}{Re} \colorbox{myellow}{Released} \colorbox{mred}{reve} \colorbox{myellow}{Game} \colorbox{myellow}{re} \colorbox{myellow}{November} \colorbox{myellow}{It} \colorbox{myellow}{in} \colorbox{myellow}{In}\\ 
OPT$_{\text{EOL+CSE}}$ & \colorbox{mgreen}{\_Game} \colorbox{mgreen}{\_March} \colorbox{myellow}{\_games} \colorbox{mred}{\_Nintendo} \colorbox{myellow}{\_game} \colorbox{mgreen}{\_Microsoft} \colorbox{mgreen}{\_PlayStation} \colorbox{myellow}{\_Games} \colorbox{myellow}{Game} \colorbox{mgreen}{\_2003}\\ 
\midrule 
LLaMA & \colorbox{mred}{<0x0A>} \colorbox{mred}{\_The} \colorbox{myellow}{\_It} \colorbox{mred}{\_A} \colorbox{myellow}{\_In} \colorbox{mred}{\_This} \colorbox{mgreen}{\_Play} \colorbox{mred}{\_An} \colorbox{mred}{\_As} \colorbox{mred}{</s>}\\ 
LLaMA$_{\text{EOL}}$ & \colorbox{myellow}{Re} \colorbox{myellow}{it} \colorbox{myellow}{re} \colorbox{myellow}{It} \colorbox{mgreen}{\_Re} \colorbox{mgreen}{\_it} \colorbox{myellow}{\_It} \colorbox{myellow}{in} \colorbox{mred}{The} \colorbox{myellow}{In}\\ 
LLaMA$_{\text{EOL+CSE}}$ & \colorbox{myellow}{\_game} \colorbox{myellow}{\_games} \colorbox{mgreen}{\_Game} \colorbox{myellow}{game} \colorbox{myellow}{Game} \colorbox{myellow}{\_Games} \colorbox{mgreen}{\_March} \colorbox{myellow}{\_release} \colorbox{mgreen}{\_released} \colorbox{mgreen}{\_November}\\ 
\midrule 
Mistral & \colorbox{mgreen}{,} \colorbox{mgreen}{\_and} \colorbox{mgreen}{2} \colorbox{mgreen}{\_} \colorbox{mred}{1} \colorbox{mgreen}{\_in} \colorbox{mred}{\_(} \colorbox{mred}{\_as} \colorbox{mred}{-} \colorbox{mred}{\_the}\\ 
GritLM & \colorbox{mgreen}{\_Game} \colorbox{mgreen}{\_Xbox} \colorbox{mred}{\_Pok} \colorbox{myellow}{\_game} \colorbox{mred}{\_cross} \colorbox{mred}{\_revealed} \colorbox{mgreen}{\_Windows} \colorbox{mgreen}{,} \colorbox{mgreen}{\_} \colorbox{mred}{\_reveal}\\ 
LLM2Vec & \colorbox{myellow}{\_release} \colorbox{myellow}{\_releases} \colorbox{mgreen}{\_released} \colorbox{myellow}{\_Release} \colorbox{mred}{\_revealed} \colorbox{myellow}{\_releasing} \colorbox{myellow}{release} \colorbox{mgreen}{\_Xbox} \colorbox{myellow}{\_game} \colorbox{mred}{\_reveal}\\ 
\bottomrule
\end{tabular}
\caption{The top 10 aligned tokens for eight $\hat{f}$ for text embedding and their corresponding $f$ for text generation when the input text is ``\textit{Revealed in March 2003, it was released across Game Boy Advance, PlayStation 2, GameCube, Xbox and Microsoft Windows in November 2003}''.}
\label{tab:decoded_token}
\end{table*}

\begin{figure*}[ht]
    \centering
    \includegraphics[width=\textwidth]{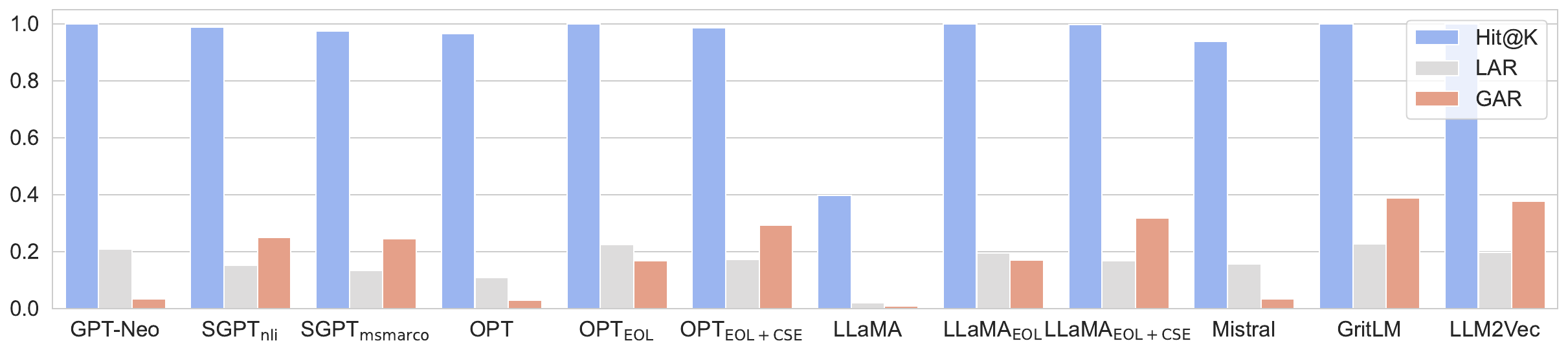}
    \caption{The comparison of evaluation metric when embedding with eight embedders $\hat{f}$ and their corresponding $f$.}
    \label{fig:evaluation_metric}
\end{figure*}

\subsection{Experiment}

\paragraph{Dataset $D$} We randomly sample 10K of the 1M Wikipedia texts provided by \citet{gao2021simcse} and report the metric calculated by this dataset. Experiments on other datasets, such as SNLI \cite{bowman2015large} and MSMARCO  \cite{nguyen2016ms}, lead to similar conclusions.

\paragraph{Triplet $(\hat{f}$, $T$, ${\bf E}_g)$} We select eight LLM-based embedders for analysis, which include SGPT$_{\text{nli}}$ and SGPT$_{\text{msmarco}}$ \cite{muennighoff2022sgpt}; OPT$_{\text{EOL}}$, OPT$_{\text{EOL+CSE}}$, LLaMA$_{\text{EOL}}$ and LLaMA$_{\text{EOL+CSE}}$ \cite{jiang2023scaling}; GritLM \cite{muennighoff2024generative} and LLM2Vec \cite{behnamghader2024llm2vec}. The key information overview of these models is placed in Table \ref{tab:model}. We consider these embedders as $\hat{f}$ and obtain $T$ and ${\bf E}_g$ from their LLM backbone. To ensure the generalizability of subsequent conclusions, the embedders selected have different architectures, fine-tuning methods, and embedding methods~\footnote{Regardless of what the similarity metric is recommended, we use a simple matrix multiplication between ${\bf E}_g$ and ${\bf h}_i$, to ensure consistency with the original decoding process.}. Note that none of these embedders goes beyond what we describe in \S\ref{sec:background}.

\subsection{Analysis of Aligned Tokens}\label{qualitative_analysis}
\paragraph{Qualitative Study} We sample an input text from $D$ and show the top 10 aligned tokens of the text embedding, i.e., $\hat{T}_{s_i}^{\text 10}$, in Table \ref{tab:decoded_token}. We also show the aligned tokens for the original $f$, using the same pooling strategy as the corresponding $\hat{f}$ for fair comparison. To indicate the relationship between each token and the surface token set $T_{s_i}$, we use different colors to mark: \colorbox{mgreen}{Green} represents that the token is in $T_{s_i}$; \colorbox{myellow}{Yellow} represents that the token and a token in $T_{s_i}$ are same after stemming or lemmatization~\footnote{We use the tools provided by NLTK \cite{loper2002nltk}: SnowballStemmer for stemming and WordNetLemmatizer for lemmatization.}; \colorbox{mred}{Red} represents that the token and all tokens in $T_{s_i}$ have no literal connection. As shown in Table \ref{tab:decoded_token}, we find that (1) the text embeddings from the original $f$ align with some tokens related $T_{s_i}$, but most of them are meaningless tokens, such as ``and'' and ``the'' etc; (2) compared to those aligned from $f$, the text embeddings from $\hat{f}$ also align with the tokens related to $T_{s_i}$ but more meaningful, such as ``game'' and ``November''; (3) even though some tokens are marked red, this only means that they are literally unrelated to $T_{s_i}$, but there may be a deeper connection. For example, ``Nintendo'' is the development company of ``Game Boy Advance'' in the input text.

\paragraph{Quantitative Study} To quantitatively reflect the connection between $\hat{T}_{s_i}^K$ and $T_{s_i}$, we propose three evaluation metrics:

\paragraph{Hit@K} To measure whether the top $K$ tokens of $\hat{T}_{s_i}$ contains any token in $T_{s_i}$, we propose the metric of Hit@K as follows:
\begin{equation}\label{eqn:hit}
    \text {Hit@K} = \underset{s_i \in D}{\mathbb{E}}\left[\mathbb{I}\left(\left|\hat{T}_{s_i}^K \cap T_{s_i}\right|>0\right)\right]
\end{equation}
where $\mathbb{I}(.)$ is the indicator function, $|\cdot|$ represents the element number of the set.

\paragraph{Local Alignment Rate} To measure the overlap degree between the tokens in $T_{s_i}$ and the top $|T_{s_i}|$ tokens in $\hat{T}_{s_i}$, we propose the metric of Local Alignment Rate (LAR) as follows:
\begin{equation}\label{eqn:localoverlap}
    \text{LAR} = \underset{s_i \in D}{\mathbb{E}}\left[\left|\hat{T}_{s_i}^{K_i} \cap T_{s_i}\right| / K_i \right]
\end{equation}
where $K_i$ is denoted as $|T_{s_i}|$ for simplicity.

\paragraph{Global Alignment Rate} LAR can not reflect the global alignment situation. For example, elements in $\hat{T}_{s_i}^{K_i} \cap T_{s_i}$ and $\hat{T}_{s_j}^{K_j} \cap T_{s_j}$ can be either the completely same or completely different, but cannot be reflected in LAR. To measure the overlap degree in the dataset $D$ globally, we propose the metric of Global Alignment Rate (GAR) as follows: 
\begin{equation}\label{eqn:globaloverlap}
    \text{GAR} = \left|\cup_{i=1}^{|D|}\left(\hat{T}_{s_i}^{K_i} \cap T_{s_i}\right)\right| / \left|\cup_{i=1}^{|D|} T_{s_i}\right|
\end{equation}
where $|D|$ represents the text number of $D$.

\begin{figure*}[tbp]
    \centering
     \subfloat[GPT-Neo $\rightarrow$ SGPT$_\text{nli}$]{\label{fig:pca_neo}\includegraphics[width=0.25\textwidth]{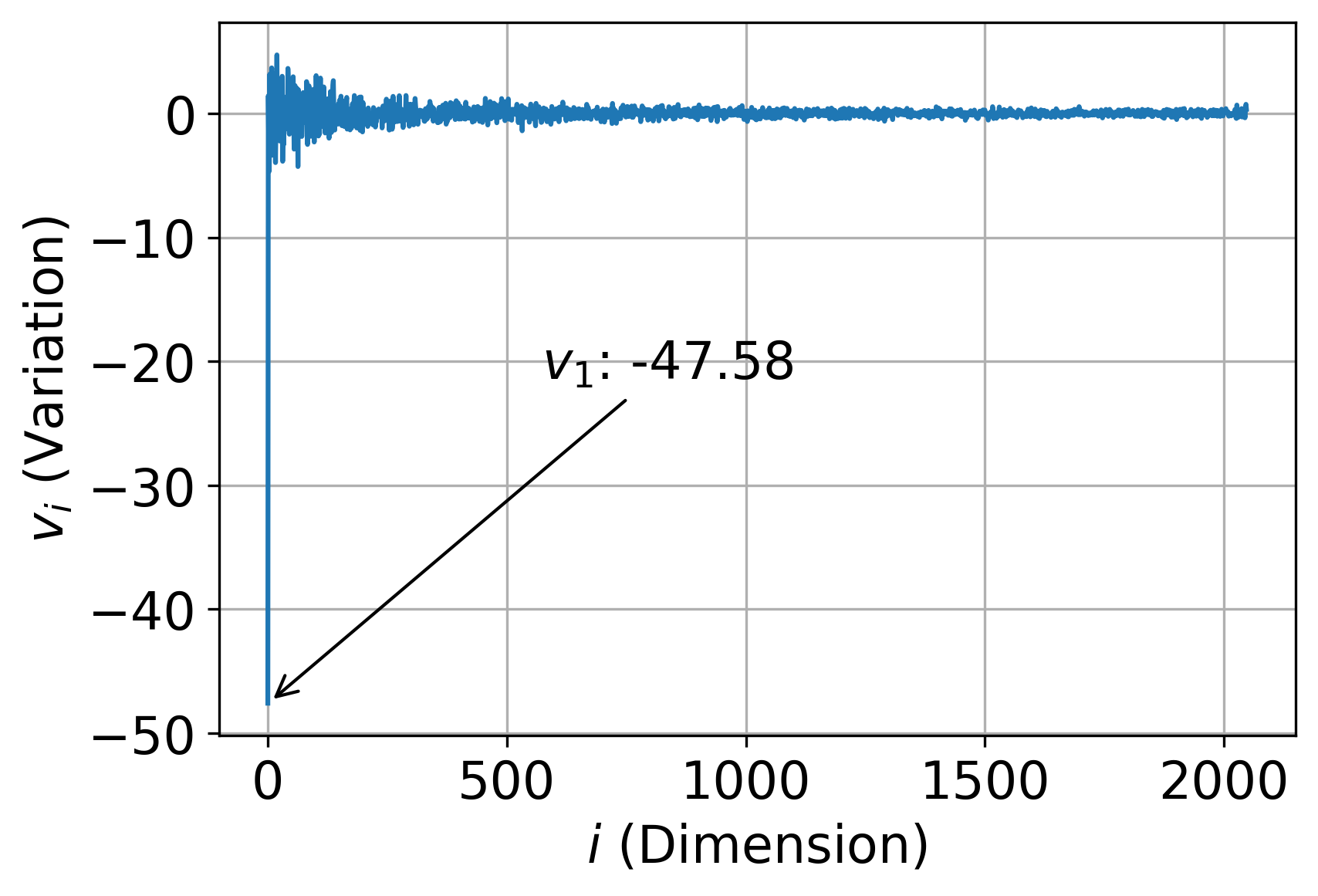}}
    \subfloat[OPT $\rightarrow$ OPT$_\text{EOL}$]{\includegraphics[width=0.25\textwidth]{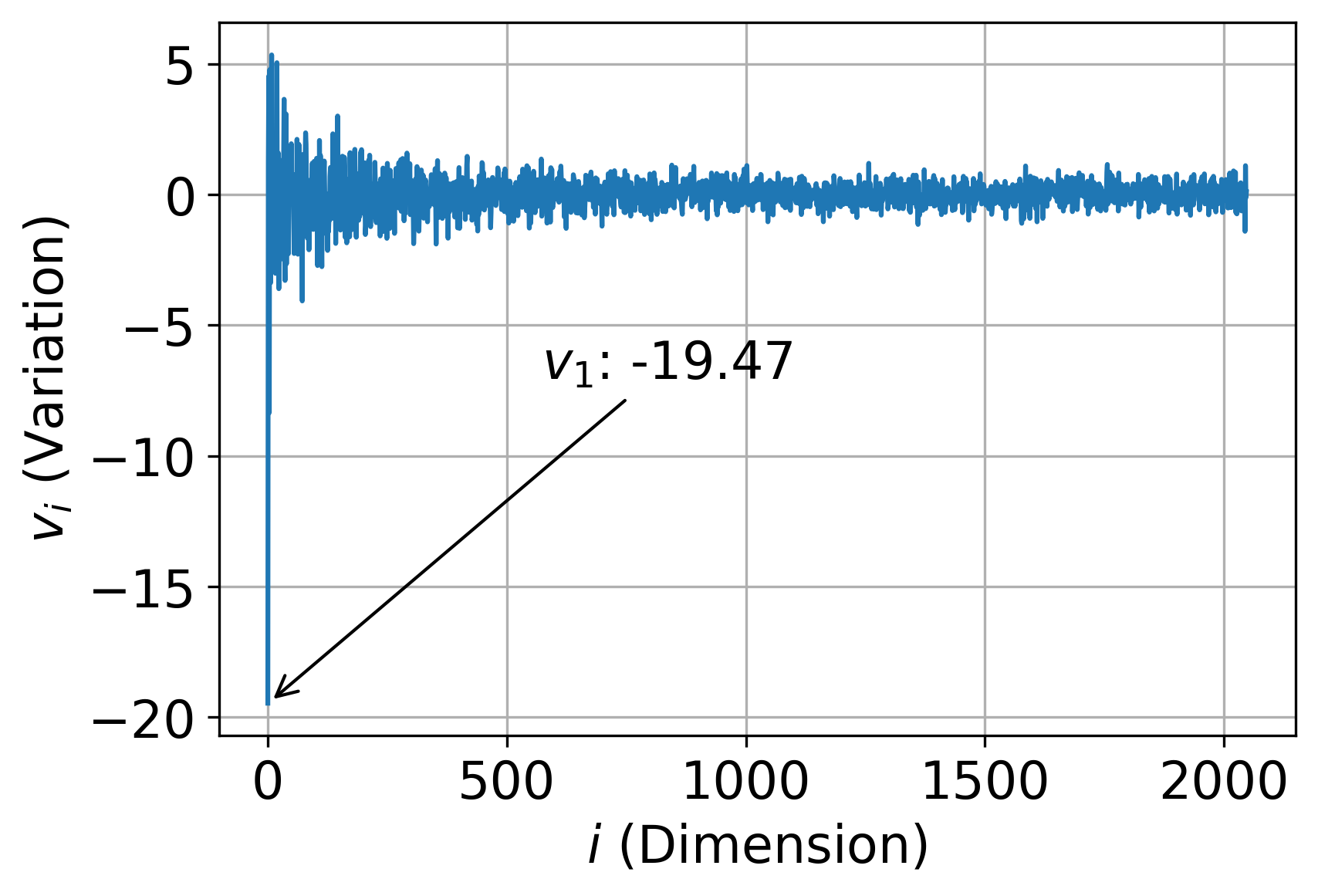}}
    \subfloat[LLaMA $\rightarrow$ LLaMA$_\text{EOL}$]{\includegraphics[width=0.25\textwidth]{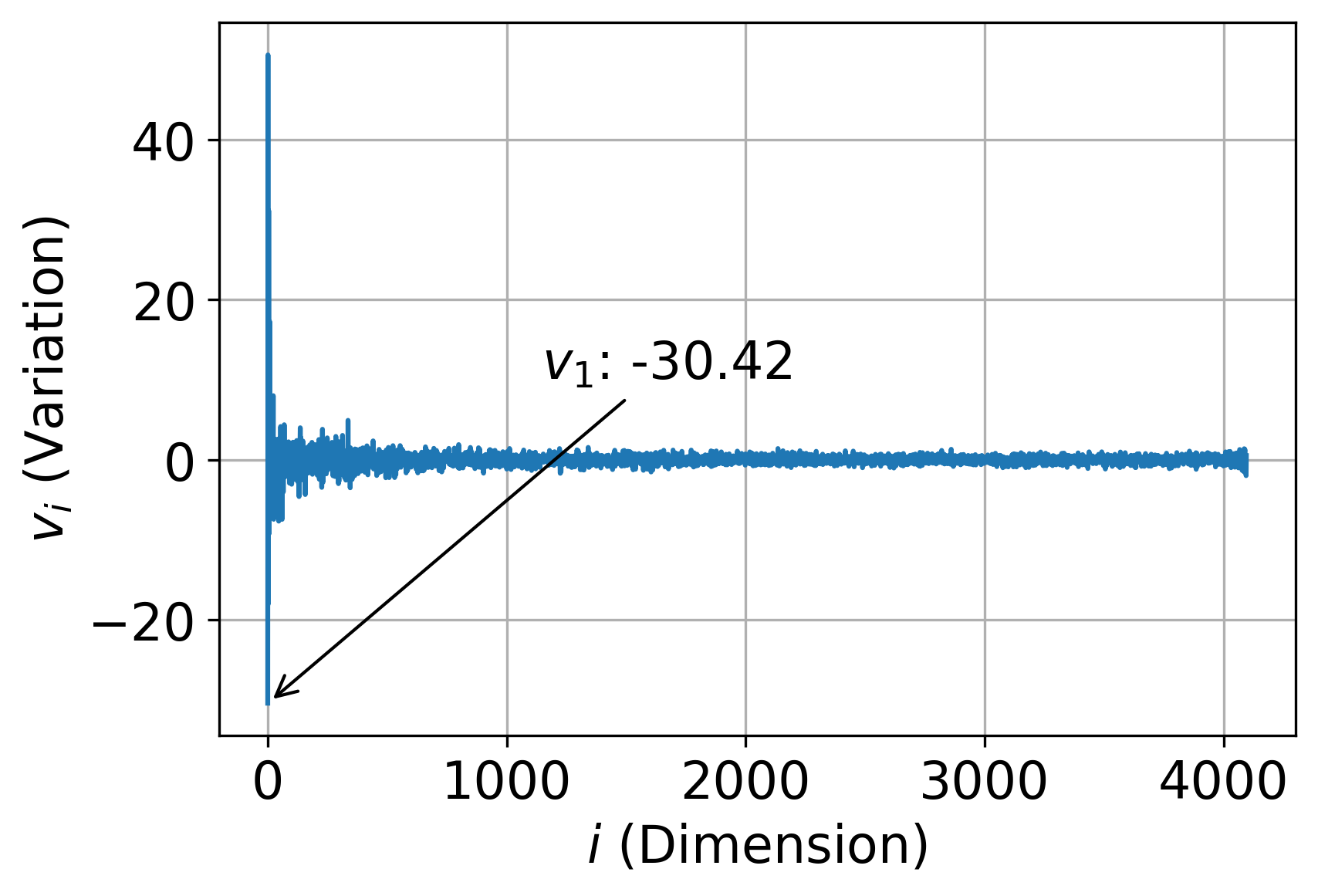}}	
    \subfloat[Mistral $\rightarrow$ LLM2Vec]{\includegraphics[width=0.25\textwidth]{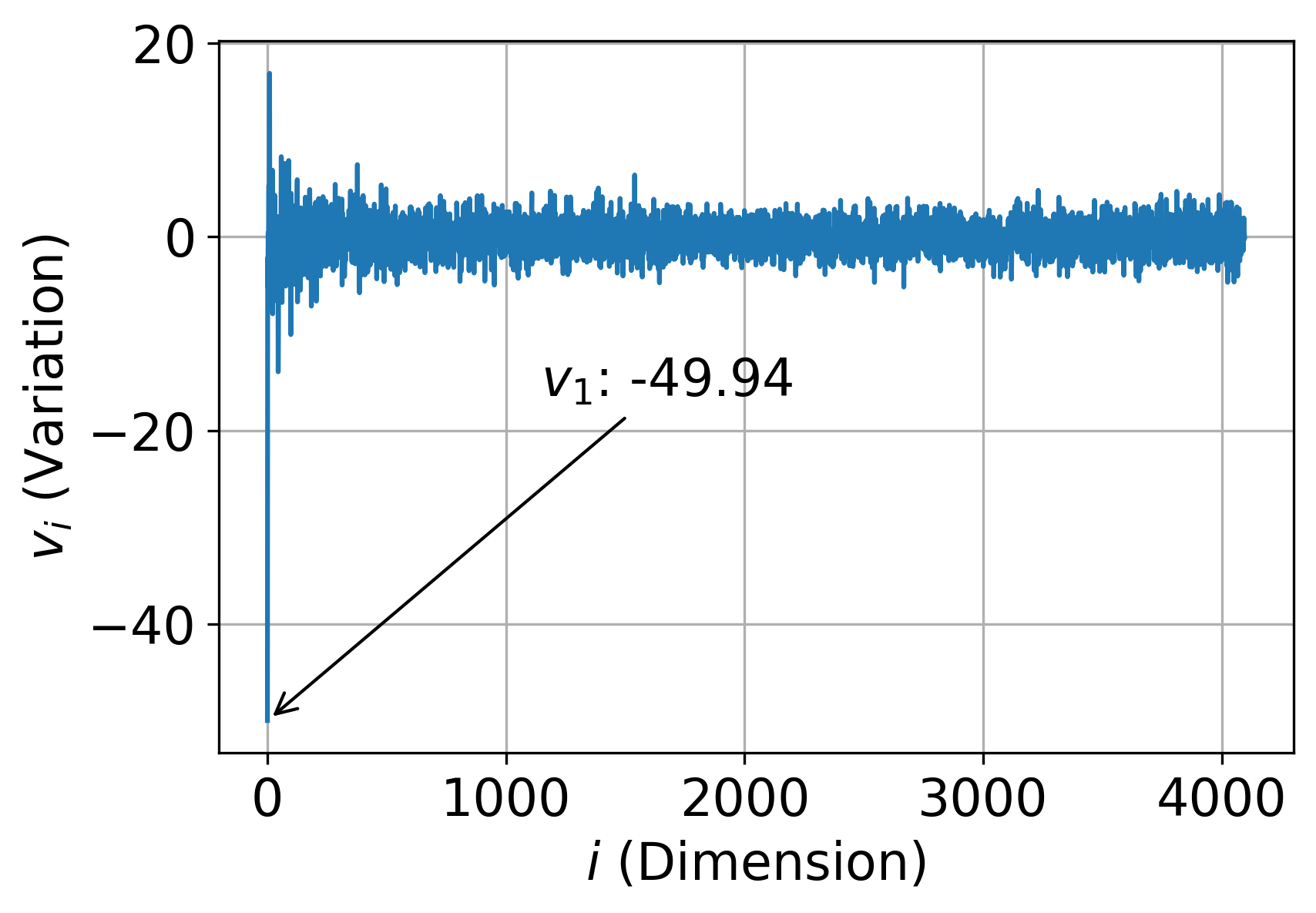}} \\
    \subfloat[GPT-Neo $\rightarrow$ SGPT$_\text{msmarco}$]{\includegraphics[width=0.25\textwidth]{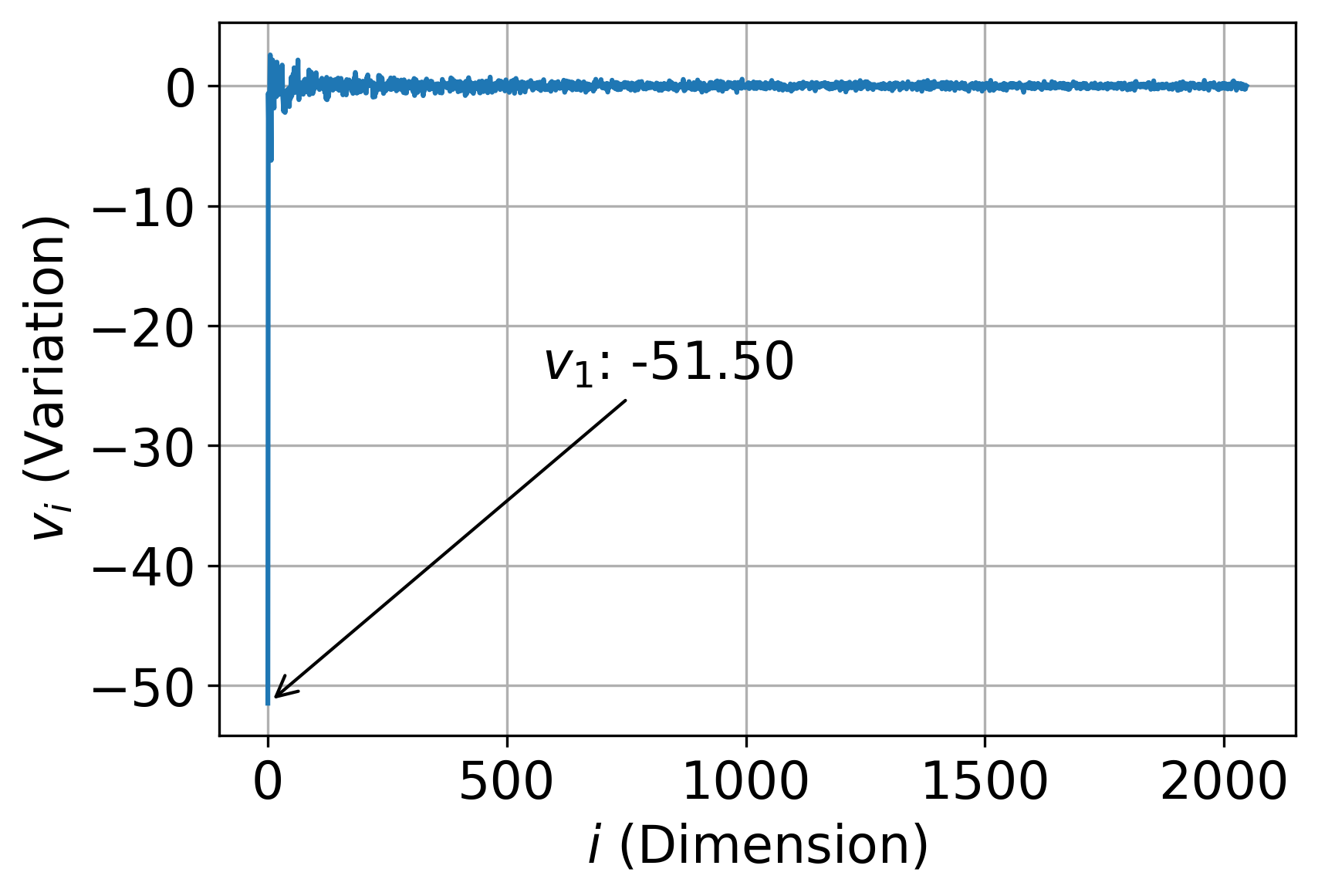}}
    \subfloat[OPT $\rightarrow$ OPT$_\text{EOL+CSE}$]{\label{fig:pca_opt}\includegraphics[width=0.25\textwidth]{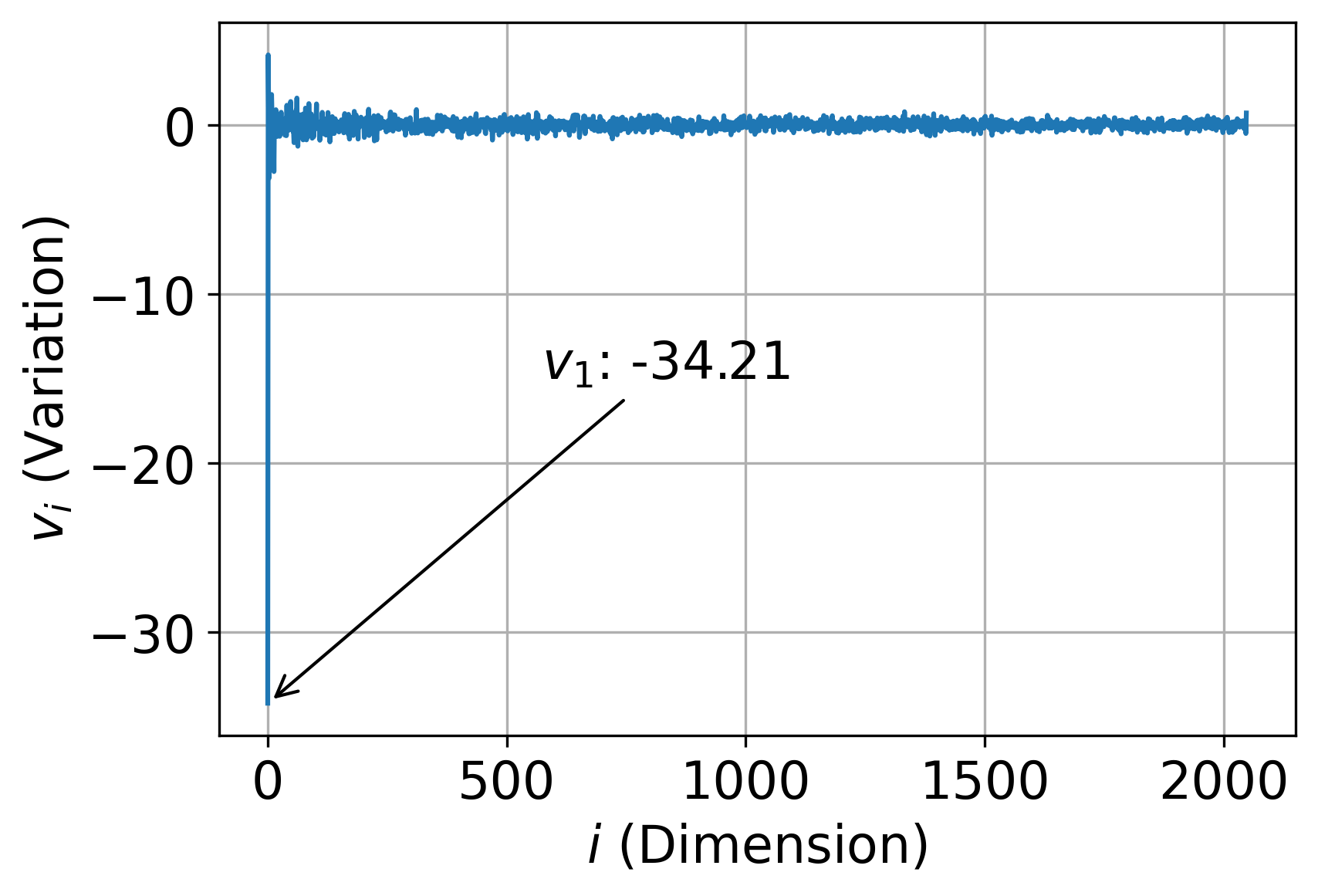}}
    \subfloat[LLaMA $\rightarrow$ LLaMA$_\text{EOL+CSE}$]{\label{fig:pca_llama}\includegraphics[width=0.25\textwidth]{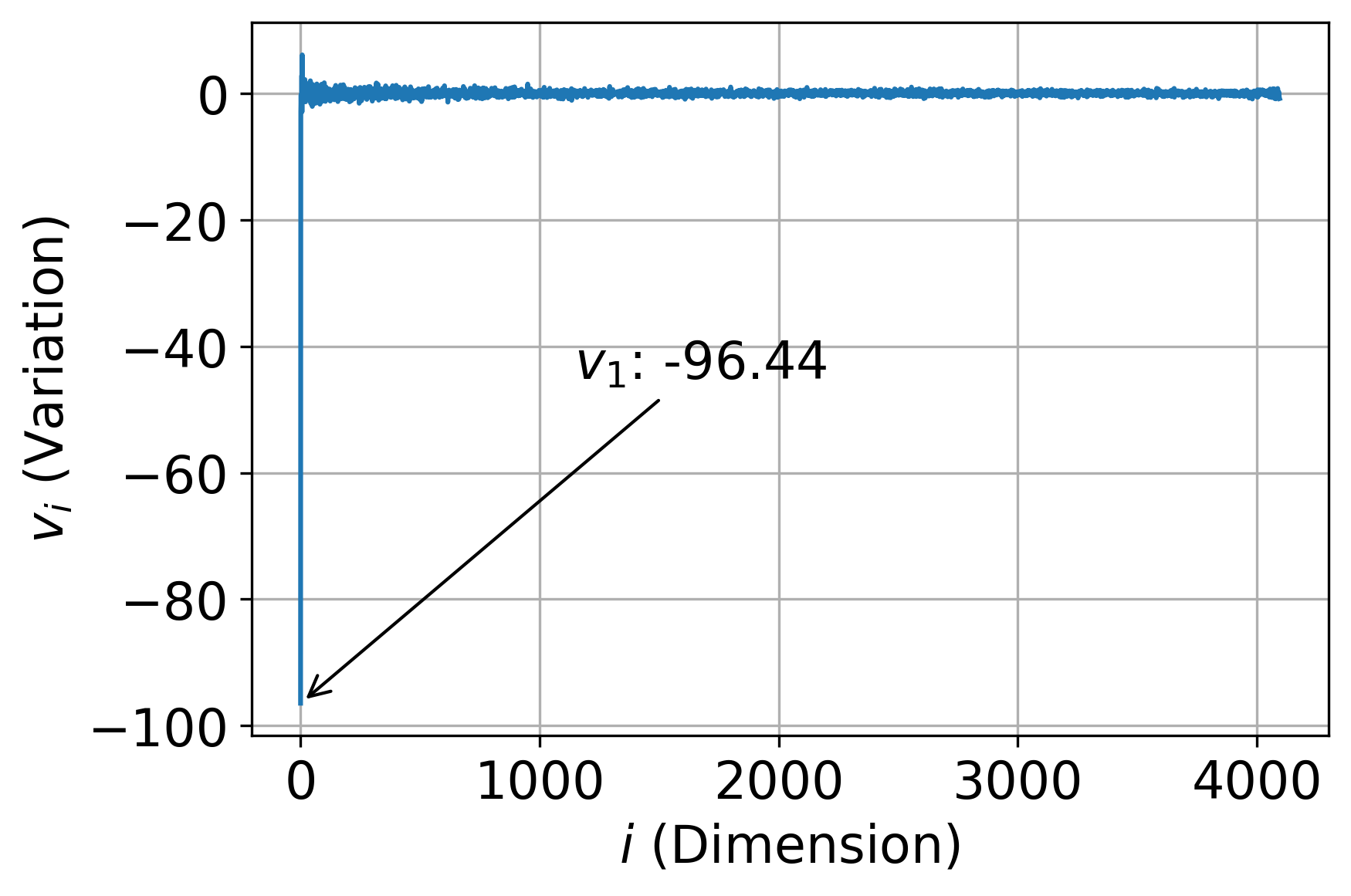}}	
    \subfloat[Mistral $\rightarrow$ GritLM]{\label{fig:pca_mistral}\includegraphics[width=0.25\textwidth]{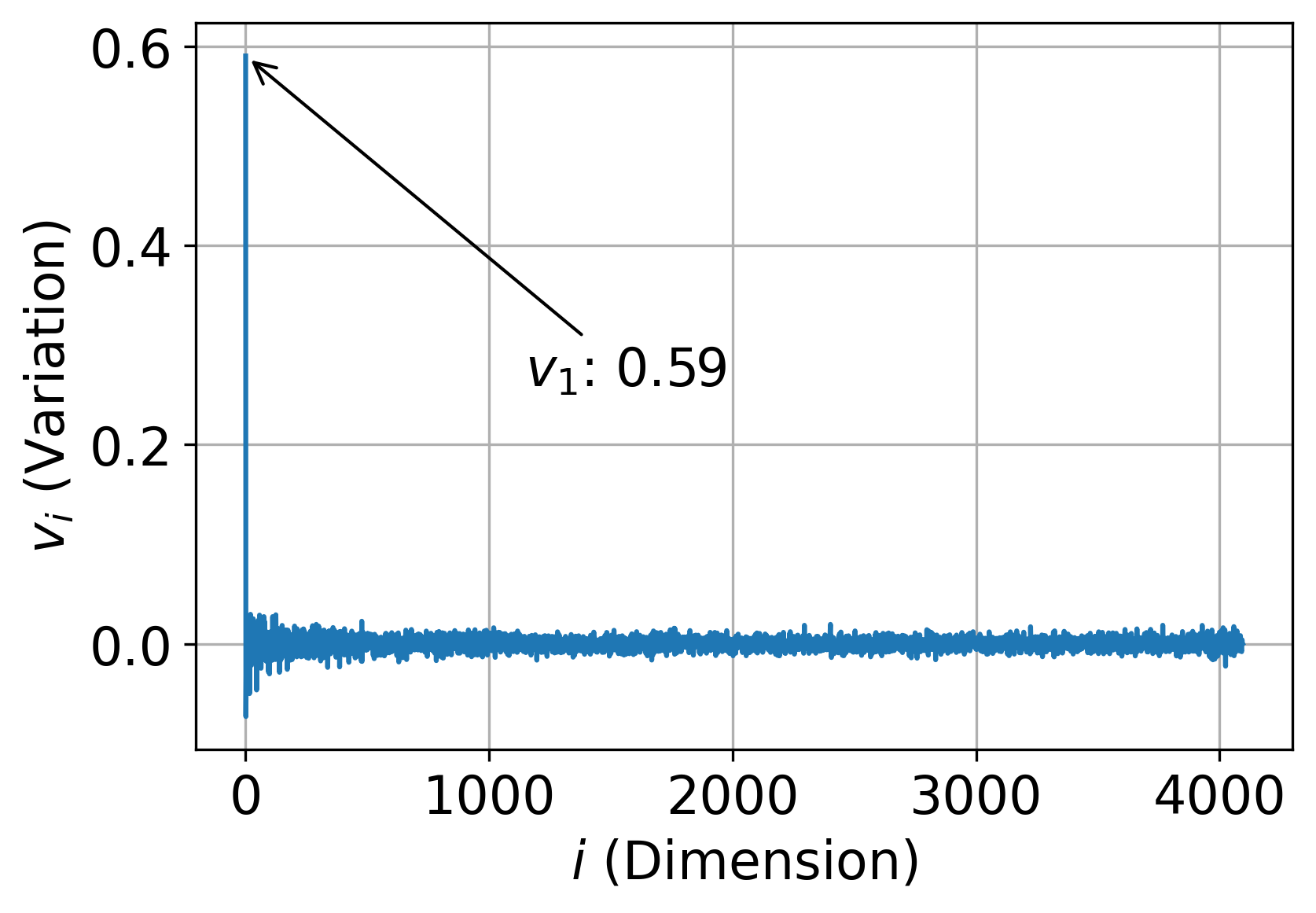}}\\	
    \caption{The variation in each principal component of the embedding space.}
    \label{fig:first_principal}
\end{figure*}

We report the Hit@10, LAR, and GAR for all embedders $\hat{f}$ and their corresponding $f$ used for text embedding in Figure \ref{fig:evaluation_metric}. The following findings can be easily concluded: (1) all $f$ and $\hat{f}$ except LLaMA maintain a high Hit@10, which means at least one token in the input text is aligned;  (2) all $\hat{f}$ also maintain a low LAR and but higher GAR than that of the corresponding $f$; (3) compared to OPT$_{\text{EOL}}$ and LLaMA$_{\text{EOL}}$, OPT$_{\text{EOL+CSE}}$ and LLaMA$_{\text{EOL+CSE}}$ lead to a lower LAR and a higher GAR after contrastive learning.

Combined with the qualitative analysis, we conclude that text embeddings from $f$ and $\hat{f} $consistently align certain tokens in the text and that $\hat{f}$-aligned tokens tend to be more diverse and more meaningful to the input text.

\section{Spectral Analysis of Embedding Space}\label{sec:analysis}

For a deeper understanding of the phenomenon, we analyze the singular value spectrum of the embedding space before and after training. Specifically, we use the same text dataset $D$ in Section \ref{sec:align} and some $(f, \hat{f})$ pairs, while all texts in $D$ are converted into embeddings via $f$ and use the SVD decomposition to obtain a set of standard orthogonal bases in $d$-dimensional space, which can be expressed as
\begin{equation}
    {\bf U} = \left[{\bf u}_1, \cdots, {\bf u}_d\right] \in \mathbb{R}^{d\times d}
\end{equation}
where ${\bf u}_j \in \mathbb{R}^{d \times 1}$ corresponds to the singular vector of $j$-th largest singular value.

For any text $s_i$ from $D$, we denote its embedding obtained from $f$ and $\hat{f}$ as ${\bf h}_i$ and $\hat{\bf h}_i$, separately. Then we metric the variation in each principal component between ${\bf h}_i$ and $\hat{\bf h}_i$ based on ${\bf U}$:

\begin{equation}
    v_j = \underset{s_i \in D}{\mathbb{E}}\left[\left(\hat{\bf h}_i - {\bf h}_i\right)^\top {\bf u}_j \right]
\end{equation}
where $v_j$ represents the variation in the $j$-th largest principal component. Due to space limitations, we select four $(f, \hat{f})$ pairs and plot their $\{v_j\}_{j=1}^d$ in Figure \ref{fig:first_principal}. Then we have the observation as belows:

\begin{observation}\label{ob:variation}
    Compared to the original embedding space, the variation of the first principal component, i.e., $v_1$, is dominant.
\end{observation}

\begin{figure*}[ht]
    \centering
    \subfloat[Contribution to the aligned tokens.]{\label{fig:pca_contribution}\includegraphics[width=0.5\textwidth]{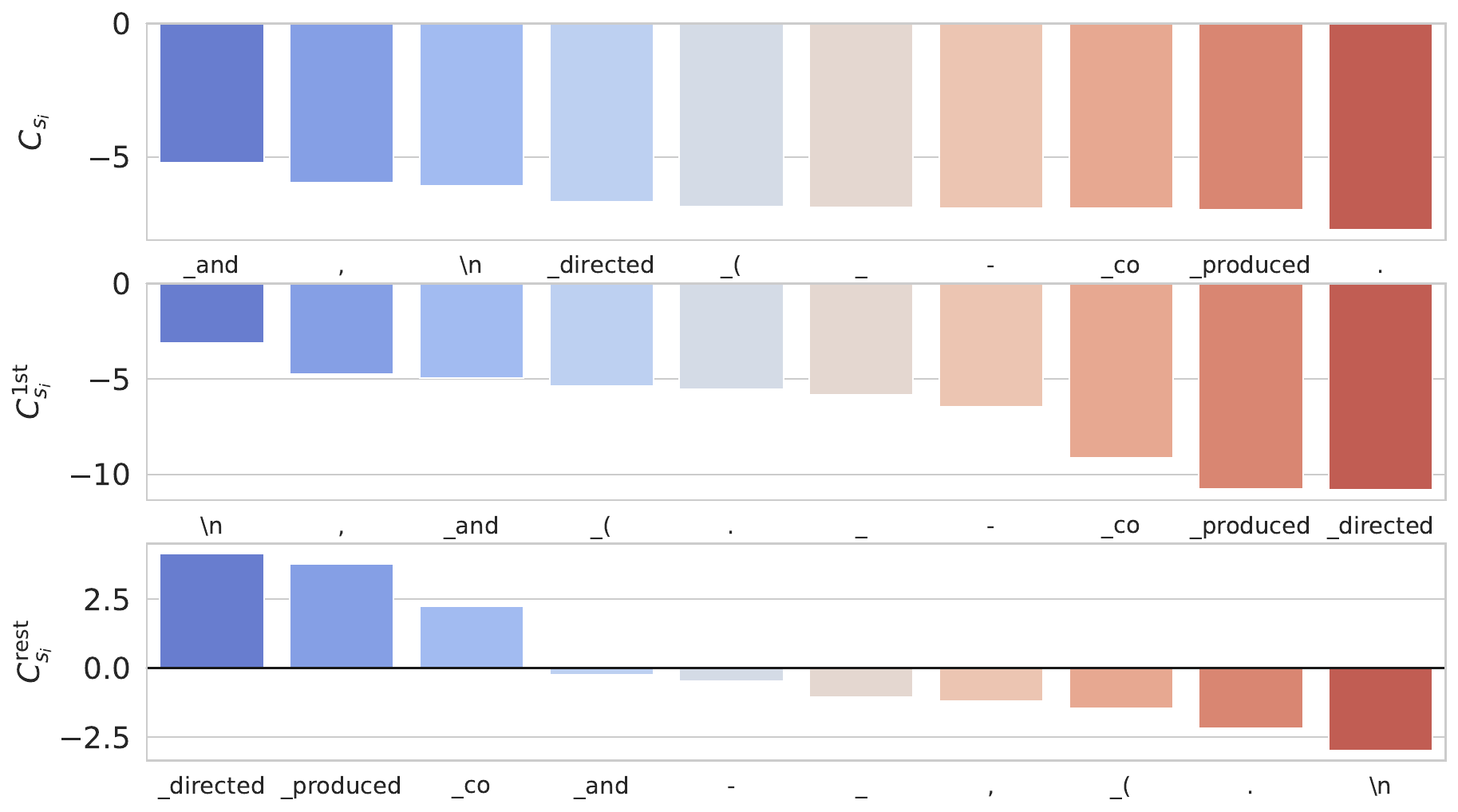}}
    \subfloat[Aligned tokens after adjusting ${\bf u}_1$.]{\label{fig:pca_adjusted}\includegraphics[width=0.5\textwidth]{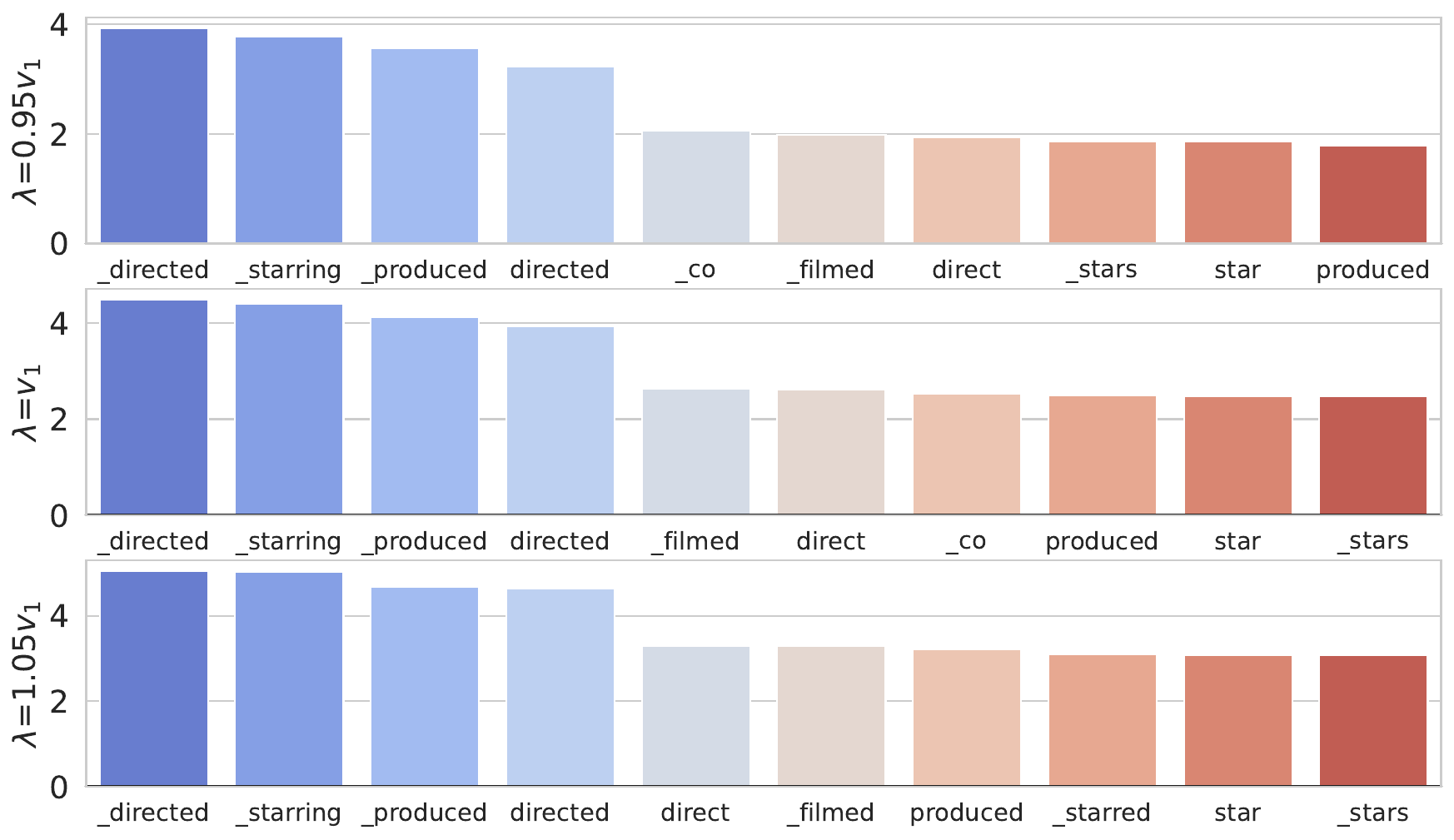}}\\	
    \caption{The situation of the aligned token when $f$ is GPT-Neo, $\hat{f}$ is SGPT$_\text{nli}$, and the input text is ``\textit{Making a Killing is a 2018 Canadian-American crime-mystery film co-written, co-produced and directed by Devin Hume.}''}
    \label{fig:first_principal_adjust_example}
\end{figure*}

Specifically, compared with the original LLMs, the embedding spaces of the most $\hat{f}$ models decrease significantly on the first principal component. Two special cases are LLaMA$_\text{EOL}$ and GritLM: (1) LLaMA$_\text{EOL}$varies greatly in each of the first few principal components. We conjecture that the anomalies of LLaMA$_\text{EOL}$ indicate precisely that its embedding space is not good enough. It is corroborated by the fact that LLaMA$_\text{EOL+CSE}$ in Figure \ref{fig:first_principal} behaves consistently with other embedders; (2) GritLM shows a small increase in the principal component. We speculate that this results from co-tuning with contrastive learning and next-token prediction. It is corroborated by the behavior of the same Mistral-based LLM2Vec, which is fine-tuned with contrastive learning only and has a decrease in the first principal component.

We further analyze the contribution of the first principal component and the other components in aligning tokens. Specifically, we divide the text embedding ${\bf h}_i$ into two components:
\begin{equation}
    {\bf h}_i = {\bf h}_i^{\text{1st}} + {\bf h}_i^{\text{rest}}
\end{equation}
where ${\bf h}_i^{\text{1st}}={\bf u}_1^\top {\bf h}_i {\bf u}_1$ and ${\bf h}_i^{\text{rest}}=\sum_{j=2}^d {\bf u}_j^\top {\bf h}_i {\bf u}_j$. We then measure the contribution of ${\bf h}_i^{\text{1st}}$ and ${\bf h}_i^{\text{rest}}$ to aligning tokens. Based on the matrix decomposition, we divide the contribution into two parts:
\begin{equation}
    \underbrace{{\bf E}_g {\bf h}_i}_{C_{s_i}} = \underbrace{{\bf E}_g {\bf h}_i^{\text{1st}}}_{C_{s_i}^{\text{1st}}} + \underbrace{{\bf E}_g {\bf h}_i^{\text{rest}}}_{C_{s_i}^\text{rest}}.
\end{equation}
Specifically, we sample a text $s_i$ from $D$, rank and obtain the top $K$ tokens based on $C_{s_i}$ and see how much $C_{s_i}^{\text{1st}}$ and $C_{s_i}^{\text{text}}$ contribute to the logits. Due In Figure \ref{fig:pca_contribution}, we provide an example and obtain the following observation:

\begin{observation}\label{ob:contribution}
    The first principal component contributes much more to meaningless tokens than meaningful tokens.
\end{observation}

Combining Observation \ref{ob:variation} and \ref{ob:contribution}, we can see: (1) current text LLM-based embedders always maximize the perturbation of the first principal component, while (2) the first principal component contributes mainly to meaningless tokens. Therefore, we give the following hypothesis:

\begin{hypothesis}\label{hyp:adjust_pc}
    The text embeddings of original LLMs have been aligned with the key tokens but are not reflected due to the affection by the first principal component.
\end{hypothesis}

To verify the hypothesis, we manually adjust the embeddings from $f$. Specifically, considering that the variation on the other principal components is small compared to the first principal component,  we can simplify as follows:

\begin{equation}
\begin{aligned}
    & \underset{s_i\in D}{\mathbb{E}}\left[\left(\hat{\bf h}_i - {\bf h}_i\right)^\top {\bf U} \right] \approx [v_1, 0,\cdots, 0] \\
    \Rightarrow & \underset{s_i\in D}{\mathbb{E}}\hat{\bf h}_i \approx  \underset{s_i\in D}{\mathbb{E}} {\bf h}_i + v_1 {\bf u}_1
\end{aligned}
\end{equation}

Therefore, for each text embedding ${\bf h}_i$, we subtracted a certain amount of the first principal component and obtained the adjusted embedding ${\bf h}_i^{\text{adj}}$:
\begin{equation}
    {\bf h}_i^{\text{adj}} = {\bf h}_i + \lambda {\bf u}_1
\end{equation}
where $\lambda \in R$ is a hyper-parameter. In Figure \ref{fig:pca_adjusted}, we report the top 10 tokens aligned by ${\bf h}_i^{\text{adj}}$ and their corresponding logits when adjusting $\lambda$ for 0.95$v_1$, $v_1$ and 1.05$v_1$. As shown in Figure \ref{fig:pca_adjusted}, the embedding from $f$ can align with more meaningful tokens of the input text by adjusting only the first principal component, verifying our hypothesis. The similar conclusions are shown on $f$ of other studies.

\section{Potential Application}\label{sec:application}

\subsection{Training-Free Embedding Sparsification}
The LLM-based embedders show superior Information Retrieval (IR) performance over the embedding models based on Transformer encoder-only PLMs (e.g., BERT \cite{kenton2019bert} and RoBERTa \cite{liu2019roberta}). However, the dimensionality of these LLMs' output embeddings (2048$\sim$4096) far exceeds the dimensionality of BERT and RoBERTa (768$\sim$1024), which will incur exponential computation and storage overhead in practice. To overcome this problem, we propose a new sparse retrieval method to generate high-quality query extensions for queries and sparse representations for documents.

For each document $d_i$, we obtain its embedding $\hat{\bf h}_{d_i}$ and aligned token set $\hat{T}_{d_i}$ using the embedding LLM. Then we can maintain a vocabulary-length sparse vector $\tilde{h}_{d_i}=[w_{t_1}, \cdots, w_{t_L}]$, where only those dimensions corresponding to the top $K$ aligned tokens are not zero:
\begin{equation}
    w_{t_i} = 
    \begin{cases}
    {\bf e}_{t_i}^\top \hat{\bf h}_{d_i} & \text{if } t_i \in \hat{T}_{d_i}^K \\
    0 & \text{otherwise}
    \end{cases}
\end{equation}

For each query $q_i$, we get its literal token set $T_{q_i}$ using the tokenizer and its aligned token set $\hat{T}_{q_i}$. It is easy to see that we can extend $T_{q_i}$ using the first $M$ elements in $\hat{T}_{q_i}$, obtaining the expanded token set $\tilde{T}_{q_i}=T_{q_i} \cup \hat{T}_{q_i}^M$.

In ad-hoc retrieval scenarios, all document sparse representations can be computed and cached in advance while the query is computed and extended on the fly. Therefore, we can calculate the similarity of $q_i$ and $d_j$ as follows:

\begin{equation}\
    \text{Similarity}(q_i, d_j) = \sum\nolimits_{t_k \in (\tilde{T}_{q_i} \cap \hat{T}_{d_i}^K )} w_{t_i}
\end{equation}

We select LLM2Vec and GritLM due to their SOTA performance but up to 4096 embedding dimensions. For evaluation, we select four information retrieval datasets: FiQA \cite{maia201818}, NFCorpus \cite{boteva2016full}, SciFact \cite{wadden2020fact} and ArguAna \cite{wachsmuth2018retrieval} and report the nDCG@10. For hyper-parameter, we experiment under the settings $K \in \{1000, 2000, 3000\}$ and $M \in \{25, 50, 75, 100\}$ and report the best results in Table \ref{tab:spa_ir}. In the most datasets, the performance is insensitive to $K$, while increasing with the increase of $M$.

\begin{table}[th]\small
    \centering
    \begin{tabular}{l|cccc}
    \toprule
        {\bf Model} & {\bf FiQA} &  {\bf NFCorpus} & {\bf SciFact} & {\bf ArguAna} \\
    \midrule
        BM25 & 0.236 & 0.325 & 0.665 & 0.315 \\
        SPLADEv2 & 0.336 & 0.334 & 0.693 & 0.479 \\
    \midrule
        LLM2Vec & 0.531 & 0.393 & 0.789 & 0.575 \\
        \rowcolor{gray!20} \quad to Spar. & 0.404 & 0.326 & 0.669 & 0.481 \\
    \midrule
        GritLM & 0.600 & 0.409 & 0.792 & 0.632 \\
        \rowcolor{gray!20} \quad to Spar. & 0.457 & 0.336 & 0.703 & 0.526 \\
    \bottomrule
    \end{tabular}
    \caption{The performance on four IR datasets. ``to Spar.'' expresses our sparse retrieval method.}
    \label{tab:spa_ir}
\end{table}

Our sparse retrieval approach preserves 80\% of the text embeddings' performance, outperforming the strong baselines: BM25 and SPLADEv2. Since the length of sparse representation is fixed, our sparse retrieval method can achieve a retrieval efficiency similar to that of BM25 when ignoring the consumption of the query encoding process. Compared to the original dense retrieval method, our method only needs $\sim$13\% FLOPs in the inference stage, with plenty of room for further improvement.
\subsection{Explain Instruction-Following Capability}
Recent works such as Instructor \cite{su2023one} and InBedder \cite{peng2024answer} use different instruction prefixes to distinguish different embedding tasks. To explain how the instruction-following embedder works, we show that the same text will align to different key tokens when prompted by the task-specific instruction. Considering a toy example of three sentences: $(S_A, S_B, S_C)$ and one instruction $I$:

\begin{flushleft}
    \textbf{$S_A$}: \textcolor{mblue}{I really enjoyed the movie last night.} \\
    \textbf{$S_B$}: \textcolor{mblue}{I didn't enjoy the movie last night at all.} \\
    \textbf{$S_C$}: \textcolor{mblue}{I had a great time watching the film this afternoon.\\}
    \textbf{$I$}: \textcolor{mblue}{Classify the emotion expressed in the given Twitter message into one of the six emotions: anger, fear, joy, love, sadness, and surprise.}
\end{flushleft}
where $I$ is introduced by \citet{wang2023improving} and used for EmotionClassification \cite{saravia2018carer}. We use LLM2Vec as the embedder and observe if aligned tokens from the same text differ with the instruction and without the instruction.

\begin{table}[tb]\small
    \centering
    \begin{tabular}{l|c}
        \toprule
        Setting & Top 5 aligned token of $S_A$ \\
        \midrule
        -wo $I$ & \_Movie \_movie \_cinema \_movies \_watched \\
        -w $I$ & \_Joy \_joy \_happiness joy \_Love \\
        \midrule
        & Top 5 aligned token of $S_B$ \\
        \midrule
        -wo $I$ & \_movie \_Movie \_movies \_cinema \_Mov \\
        -w $I$ & \_sad \_Sad \_disappointment \_disappointed \_anger  \\
        \midrule
        & Top 5 aligned token of $S_C$ \\
        \midrule
        -wo $I$ & \_afternoon \_cinema \_movie \_Movie \_movies \\
        -w $I$ & \_joy \_Joy joy \_happiness \_delight \\
        \bottomrule
    \end{tabular}
    \caption{Comparison of the aligned tokens with / without the instructions prefix.}
    \label{tab:different_instruction}
\end{table}

As shown in Table \ref{tab:different_instruction}, the tokens aligned by all sentences are largely changed when adding $I$. When $I$ is not added, all tokens are aligned to the non-sentiment tokens. Interestingly, when $I$ is added, $S_A$ and $S_C$ are mainly aligned to the tokens for positive emotions, while $S_B$ is mainly aligned to the tokens for negative emotions. We also find the similarities among these sentences will be different:
\begin{itemize}
    \item When no instruction is added, the embedder can only ``randomly'' select some key tokens to align. For all sentences, the LLM happens to both choose topic-related tokens. As a result, similarity $(S_A, S_B)$=\texttt{0.821} is higher than similarity $(S_A, S_C)$=\texttt{0.718}.
    \item When the instruction for sentiment classification is added, the LLM ``adaptively'' selects the sentiment tokens to align with. As a result, similarity $(I+S_A, I+S_B)$=\texttt{0.814} become lower than similarity $(I+S_A, I+S_C)$=\texttt{0.829}.
\end{itemize}

\subsection{Explain Semantic Relatedness / Similarity}

Text embedders are fine-tuned with different datasets depending on their evaluation task. For example, the NLI datasets are often used for training when evaluating the Semantic Text Similarity (STS) task on ``semantic similarity''. Instead, the MS MARCO dataset is often used for training when evaluating the information retrieval task on ``semantic relatedness''. It is difficult to distinguish these two fuzzy concepts for a long time \cite{abdalla2023makes}. Benefiting from our finding, we can intuitively understand ``semantic similarity'' and ``semantic relatedness'' by mapping the text embeddings to token space.
Considering a toy example of two sentences $(S_A, S_B)$:
\begin{flushleft}
    {\bf $S_A$}: \textcolor{mblue}{I like apples.} \quad
    {\bf $S_B$}: \textcolor{mblue}{I dislike apples.}
\end{flushleft}
We obtain the two sentence embeddings with SGPT$_{\text{nli}}$ and SGPT$_{\text{msmarco}}$ and obtain the aligned tokens with the decoder layer of GPT-Neo. As there is no difference between these two embedders except for the fine-tuning dataset.

As shown in Table \ref{tab:different_data}, most aligned tokens of $S_A$ are related to ``apple'', while there is some difference in the tokens aligned by $S_B$. Specifically, when SGPT$_{\text{nli}}$ is used, tokens related to ``dislike'' are in the majority, whereas when SGPT$_{\text{msmarco}}$ is used, the ratio of tokens related to ``dislike'' and ``apple'' is balanced. This difference can help intuitively understand the difference between ``semantic similarity'' and ``semantic relatedness'':
\begin{itemize}
    \item $S_A$ and $S_B$ are not considered to have a high degree of similarity because $S_A$ is an affirmative while $S_B$ is a negative sentence. SGPT$_{\text{nli}}$ aligns the embedding of $S_B$ to ``dislike'' to ensure that the embedding of the two sentences is far enough apart. Therefore, the cosine similarity given by SGPT$_{\text{nli}}$ is only \texttt{0.419};
    \item $S_A$ and $S_B$ can be considered highly relevant because they both describe whether ``I'' like ``apples'' or not. SGPT$_{\text{msmarco}}$ aligns the embedding of $S_B$ to both "dislike" and "apple" to ensure that the final similarity reflects their relevance. Therefore, the cosine similarity given by SGPT$_{\text{msmarco}}$ is \texttt{0.816};
\end{itemize}

\begin{table}[tb]\small
    \centering
    \begin{tabular}{c|c}
        \toprule
        Model & Top 5 aligned token of $S_A$ \\
        \midrule
        SGPT$_{\text{nli}}$ & \_apple \_apples \_Apple apple Apple \\
        SGPT$_{\text{msmarco}}$ & \_apple \_Apple Apple apple \_liking \\
        \midrule
        & Top 5 aligned token of $S_B$ \\
        \midrule
        SGPT$_{\text{nli}}$ & \_dislike \_disliked hate \_hates \_apple \\
        SGPT$_{\text{msmarco}}$ & \_dislike \_Apple \_disliked \_apple Apple \\
        \bottomrule
    \end{tabular}
    \caption{Comparison of the aligned tokens when using different fine-tuning data.}
    \label{tab:different_data}
\end{table}

\section{Related Works}
Reconstructing the information of the original text from its embedding \cite{pan2020privacy} has been explored primarily as a topic in privacy and security. Recently, some works have tried reconstructing the original text from text embeddings by training additional decoders. \citet{li2023sentence} is the first to try a single-round reduction method, while \citet{morris2023text} and \citet{chen2024text} use an iterative multi-round method, Vec2Text, to achieve better text reconstruction performance. Unlike these methods, this work does not involve any training process but only draw on the decoding layers in the LLMs.

The most related work is \citet{ram2023you}, who find that embeddings from several BERT-based models align with key tokens after passing through the MLM head from the original BERT. Our work differs in three aspects: (1) \citet{ram2023you} observe this in three models, while we find that many <1B models (e.g., SimCSE \cite{gao2021simcse}, Contriever \cite{izacard2022unsupervised} and E5 \cite{wang2022text}) do not exhibit this, motivating our focus on LLMs where the phenomenon consistently holds; (2) they describe the effect, whereas we further explain its cause via spectral analysis; (3) they focus on dense retrieval, while we extend to sparse retrieval and interpretability applications.

\section{Conclusion}
In this work, we show the alignment of text embeddings obtained from LLMs for embedding with key tokens in the input text. We first perform qualitative and quantitative analyses on eight LLMs to demonstrate the generalizability of our conclusions. Then, we use spectral analysis to understand the phenomenon better and show that text embeddings can be aligned to key tokens by adjusting the first principal component. For application, three examples given on information retrieval and interpretability demonstrate our findings' broad application promise and continued research value.

\section*{Limitation}
We summarize the limitations as follows:

\begin{itemize}
    \item For universality, we cannot observe a similar phenomenon in the encoder-only PLM-based embedders except for several special cases. We conjecture that the reason comes from two sources: (1) encoder-only PLMs have a larger variation in the embedding space than LLMs due to too few parameters; (2) encoder-only PLMs use a complex MLM head for training, and the text embedding is obtained too far away from the final decoded token embedding matrix, resulting in no dependencies between them.
    \item For the LLM-based embedders, we only conducted the empirical study for the LLMs for English embedding. We have not extended the study to a multi-lingual setting due to insufficient LLMs for multi-lingual embedding.
    \item In Section \ref{sec:analysis}, we have only shown that adjusting the first principal component can achieve alignment with key tokens, but we are unable to explain why the LLMs’ pre-training phase leads to such an embedding space, nor can we achieve the same performance as the existing methods by tuning only the first principal component. At the same time, it is conceivable that we cannot achieve a similar embedding quality to contrastive learning by adjusting only the first principal component.
\end{itemize}

\section*{Acknowledgements}
This work was supported by the National Science and Technology Major Project under Grant 2022ZD0120202, in part by the National Natural Science Foundation of China (No. U23B2056), in part by the Fundamental Research Funds for the Central Universities, and in part by the State Key Laboratory of Complex \& Critical Software Environment.
\bibliography{custom}

\end{document}